\pdfoutput=1

\documentclass[11pt]{article}

\usepackage[final]{acl}

\usepackage{times}
\usepackage{latexsym}
\usepackage{tabularx}
\usepackage{booktabs}

\usepackage{CJKutf8}

\usepackage[T1]{fontenc}

\usepackage[utf8]{inputenc}
\usepackage{graphicx}
\usepackage{microtype}
\usepackage{float}

\usepackage{inconsolata}

\usepackage{tabularx}
\usepackage{geometry}
\usepackage{CJKutf8}

\title{Pretraining and Updates of Domain-Specific LLM: \\A Case Study in the Japanese Business Domain}

\author{Kosuke Takahashi, Takahiro Omi, Kosuke Arima \\
Stockmark \\
\texttt{\{kosuke.takahashi, takahiro.omi, kosuke.arima\}@stockmark.co.jp} \\
\AND
Tatsuya Ishigaki \\
National Institute of Advanced Industrial Science and Technology 
\\
\texttt{ishigaki.tatsuya@aist.go.jp} \\
}

\begin{document}
\maketitle
\begin{abstract}
The development of Large Language Models (LLMs) in various languages has been advancing, but the combination of non-English languages with domain-specific contexts remains underexplored.
This paper presents our findings from training and evaluating a Japanese business domain-specific LLM designed to better understand business-related documents, such as the news on current affairs, technical reports, and patents. 
Additionally, LLMs in this domain require regular updates to incorporate the most recent knowledge.
Therefore, we also report our findings from the first experiments and evaluations involving updates to this LLM using the latest article data, which is an important problem setting that has not been addressed in previous research.
From our experiments on a newly created benchmark dataset for question answering in the target domain, we found that (1) our pretrained model improves QA accuracy without losing general knowledge, and (2) a proper mixture of the latest and older texts in the training data for the update is necessary.
Our pretrained model and business domain benchmark are publicly available~\footnote{\url{https://huggingface.co/stockmark}} to support further studies.
\end{abstract}


\section{Introduction}
The development of Large Language Models (LLMs) has seen significant progress across various languages.
However, the combination of language-specific and domain-specific contexts remains underexplored.
This study focuses on a Japanese LLM tailored for the business domain, addressing the need for models that can understand and process business-related documents such as articles on current affairs, corporate activities, and social issues.

\begin{table}
    \centering
    \footnotesize
    \begin{tabular}{ll}
    \hline
Category & Question \\ \hline
Current Affairs &
\begin{tabular}{l}
{\it Which country joined NATO in }\\ {\it April 2023 in response to}
\\{\it Russia's  invasion of Ukraine?}
\end{tabular}
\\ \hline
Corporate Activities &
\begin{tabular}{l}
{\it Which Japanese startup has}\\ {\it been developing perovskite  }\\{\it  solar cells since 2022?}
\end{tabular}
\\ \hline
Social Issues &
\begin{tabular}{l}
{\it What is carbon neutrality?}
\end{tabular}\\ \hline
Trends &
\begin{tabular}{l}
{\it What is a dark store?}
\end{tabular}
\\ \hline
    \end{tabular}
    \caption{Examples of the business related questions translated from the original Japanese.}
    \label{tab:samples_bench}
\end{table}

As demonstrated by the examples of business-related questions in Table \ref{tab:samples_bench}, accurately answering these questions requires specialized knowledge about current events, corporate activities, and social issues.
Despite the success of LLMs in various language tasks, most models are pretrained on datasets consisting primarily of general English data. 
Consequently, their performance in other languages, including Japanese, and in domain-specific tasks may remain limited.

Our research aligns with two key directions in LLM development: language-specific models and domain-specific models.
Language-specific pretrained models have been developed for several languages~\citep{zhang2021cpm,german_gpt,swallow}, while domain-specific models have excelled in areas such as finance~\citep{workshop2023bloom}. 
Combining these aspects has been shown promising since the traditional advent of domain-specific BERTs~\citep{beltagy-etal-2019-scibert,ishigaki2023pretraining}.
However, many non-English languages, including Japanese, lack such combinations, especially for recent decoder-only architectures like GPTs~\citep{NEURIPS2020_1457c0d6-gpt}.
Addressing this gap requires developing fundamental domain-specific GPT.
As a case study, we present the first Japanese LLM with 13 billion parameters specifically for the business domain.

LLMs in the business domain needs to be updated regularly to address the latest business-related queries.
For example, LLMs pretrained before 2023 do not know results of the Olympic games in 2024.
While the importance of continual updates, there has been limited research on how we can better update LLMs with the latest knowledge without losing the general knowledge.
This paper focuses on a research question about the data used for updates: how can we mix recent texts with general texts to ensure that the LLM incorporates up-to-date information while still retaining general knowledge?
To address this gap, we continually pretrain our LLM with recent business documents with various ways of the mixture, ensuring that its knowledge remains current.
This approach aims to improve the model’s accuracy in reflecting the latest business trends and information.


Evaluating a domain-specific model presents unique challenges.
In addition to using general-domain benchmarks such as lm-evaluation-harness~\citep{eval-harness}, we introduce a new business-specific benchmark. 
This benchmark consists of business questions across three tasks whose inputs are 1) question-only, 2) question with automatically retrieved context, and 3) question with manually retrieved context.
Manual evaluations on this benchmark reveal that our pretrained model outperforms existing general-domain Japanese LLMs in accuracy, particularly in the question-only setting. 
Furthermore, models updated with the latest knowledge show improved accuracy in answering questions about recent events.

Our contributions are threefold: (1) we pretrain the first and largest Japanese business domain-specific LLM; (2) we present the first experiments on LLMs updated with the latest business documents and demonstrate that a proper mixture of recent and older texts in the training data is necessary; (3) we create a new benchmark with questions designed for three distinct tasks; and (4) we demonstrate the effectiveness of our pretrained model in domain-specific tasks.
This study establishes a foundational environment for comparing future language- and domain-specific LLMs and provides valuable insights for researchers working with LLMs in other domains and languages.
Our model and benchmark questions are publicly available~\footnote{\url{https://huggingface.co/stockmark/stockmark-13b}\\ \url{https://huggingface.co/datasets/stockmark/business-questions}}.

\section{Methods}

Our approach consists of three steps: collecting datasets, filtering them, and pretraining an LLM from scratch.
The trained LLM is later used for updating.

\subsection{Dataset}

To construct our pretraining dataset, we collected 19.8\% domain-specific texts and 80.2\% general-domain texts in Japanese.

For the domain-specific data, we created our original “Curated Business Corpus” and also used patent documents from the Japan Patent Office. These two corpora are designed to provide the business knowledge and technical terminology necessary for the target domain. The Curated Business Corpus was built by curating publicly available web pages published up to September 2023. We identified relevant pages using predefined URLs and a list of cue words, extracting pages that matched the URL patterns or contained at least one of the keywords. The URLs and cue words were selected to cover various aspects of the business domain, including chemistry, materials, biology, engineering, economics, current affairs, and social trends.


To ensure a diverse training dataset, we also included general-domain data from sources such as Wikipedia, CC100, mC4, and Common Crawl. These datasets provide essential general knowledge, which is crucial for handling a wide range of natural language processing tasks. Most other LLMs utilize similar general-domain datasets, though often with different filtering strategies.

\subsection{Filtering}

Filtering is essential for enhancing dataset quality.
We implemented a three-step pipeline: language identification, removal of noisy characters, and deduplication.

\noindent
\textbf{Language Identification: } Language identification is crucial for language-specific datasets.
We used a two-method pipeline to identify non-Japanese documents: a library-based method and a language characteristics-based method.
Initially, we employed the ''xlm-roberta-base-language-detection'' library~\footnote{\url{https://huggingface.co/papluca/xlm-roberta-base-language-detection}}.
For documents with uncertain results from this tool, we applied the franc library~\footnote{\url{https://github.com/wooorm/franc}}.
Any texts not recognized as Japanese were removed from the dataset.

\noindent
\textbf{Noise Character Removal: } Low-quality Japanese texts often lack proper sentence structure.
Noisy texts may consist only of dates, HTML tags of menu bars, URLs, or lack end-of-sentence punctuation marks, such as ``\begin{CJK}{UTF8}{ipxg}{。}\end{CJK}'' (Japanese period).
We removed such texts if they were deemed non-sentential to ensure the dataset’s quality and to focus the model on relevant linguistic features. 
Additionally, because English sentences end with a period (``.''), using punctuation as a clue helps in refining English sentences.
Our characteristics-based method, while simple, is effective in distinguishing languages with specific features, such as Thai which does not use punctuation.

\noindent
\textbf{Deduplication: } Finally, we deduplicated the collected documents and sentences to eliminate identical entries.
At the document level, we used Python’s built-in hash function to convert documents into hashed values, which allowed us to remove duplicates efficiently.
At the sentence level, we counted the frequency of sentences and removed those appearing more than 15 times.
Both document-level and sentence-level deduplication were performed using exact matching.
This step prevents the model from fixating on repeated data and ensures dataset diversity.

Following these preprocessing steps, our dataset comprised a total of 220 billion tokens, as detailed in Table \ref{tab:dataset_proportion}.

\subsection{Pretraining}

We used the Llama2 architecture~\citep{touvron2023llama} with 13 billion parameters for our model.
Our hyperparameters were aligned with those reported in the Llama2 paper. 
Instead of further training Meta’s Llama2 weights, our model was pretrained from scratch.

To balance the training data for each epoch, we adopted the weighting strategy used by Llama2, but doubled the amount of Wikipedia data for two main reasons.
First, Wikipedia data is relatively clean.
Second, it contains a wealth of content relevant to the business domain. 
Additionally, we doubled the size of our Curated Business Corpus to enhance the integration of domain-specific knowledge.
By increasing the proportion of clean, domain-specific data, we aimed to mitigate the impact of noisy data sources like mC4 and Common Crawl.

For infrastructure, we utilized AWS’s Trainium, a hardware accelerator specifically designed for high-performance machine learning computations.
We deployed our training scripts on 16 trn1.32xlarge instances, each equipped with Trainium, to create a robust distributed learning environment.
The distributed learning process was managed using the neuronx-nemo-megatron library~\footnote{\url{https://github.com/aws-neuron/neuronx-nemo-megatron}}, which is available on AWS’s custom accelerators.
The pretraining phase spanned 30 days to complete one epoch of training data.

\begin{table}
    \centering
    \footnotesize
    \begin{tabular}{lc}
    \hline
Dataset & Num. of tokens [billion]  \\ \hline
Curated business corpus &	9.1 \\
Patent	& 34.8 \\
Wikipedia & 1.0 \\
CC100 & 10.9 \\
mC4 & 53.2 \\
Common Crawl & 112.9 \\ \hline
    \end{tabular}
    \caption{The size of preprocessed dataset used for pretraining.}
    \label{tab:dataset_proportion}
\end{table}

\begin{table}
    \centering
    \footnotesize
    \begin{tabular}{llll}
    \hline
Dataset & Num. of examples & Language  \\ \hline
Dolly & 15,015 & translated Japanese\\
OASST & 88,838 & translated Japanese\\
Alpaca & 51,716 & translated Japanese\\
Ichikara & 10,329 & Human-authored Japanese \\\hline
    \end{tabular}
    \caption{Candidates of datasets for instruction tuning.}
    \label{tab:dataset_explanation}
\end{table}

\subsection{Updating the Pretrained Model with Latest Business Documents}
In real-world applications, such as domain-specific question answering (QA), LLMs must be updated to incorporate the most recent knowledge. The pretrained model discussed in the previous sections was trained on texts published up until September 2023. Consequently, it may struggle to answer questions about events occurring after October 2023.

To address this limitation, we continued training our pretrained model with the latest business documents published in October and November 2023. However, we are mindful of the issue of “catastrophic forgetting”~\citep{FRENCH1999128}, where acquiring new knowledge can unintentionally displace previously learned information.

To mitigate catastrophic forgetting, we employ a strategy that blends the latest business documents with randomly selected, non-latest documents from the Curated Business Corpus. This approach is inspired by previous research~\citep{scialom-etal-2022-fine}. We introduce a hyperparameter $r$, which represents the proportion of instances sampled from the non-latest document set. For instance, if $r$ is set to 0.3, then 30\% of the continual update data comes from the Curated Business Corpus.

Our experiments in the next sections show how different values of $r$ affect the performance of domain-specific QA and analyze the extent of catastrophic forgetting.

\section{Experiments}
To evaluate our LLM, we have created a benchmark, Business Question Benchmark, for business domain question answering.
We then compared our model against various Japanese LLMs using this benchmark, as well as common Japanese benchmarks, lm-evaluation-harness.

\subsection{Benchmarks}


\noindent
\textbf{Business Question Benchmark:} This benchmark consists of 50 questions written in natural language, each paired with relevant articles. The questions cover a range of topics, including recent events, company activities, social issues, and business trends. Each question is associated with web pages retrieved through automatic and manual methods.

The benchmark offers three QA settings; 1) \textbf{NoContext-QA}: In this setting, no web pages are provided as context. This allows us to assess the LLMs’ ability to generate answers based solely on their internal knowledge, without external information, 2) \textbf{AutoRAG-QA}: for this task, we use a search engine to find the most relevant web pages for each question. The highest-ranked page with available body text is selected as the context. This setting helps evaluate how well LLMs can generate answers considering both their existing knowledge and potentially irrelevant information from the web page, 3) \textbf{ManualRAG-QA}: Here, we manually select a web page that contains answers to the question. The RAG tasks are more about comprehension, requiring the model to understand and extract information from a specific page rather than relying solely on its internal knowledge.

For AutoRAG-QA and ManualRAG-QA, the text from each web page is truncated to 1000 characters. The prompts used for these tasks are detailed in Appendix \ref{fig:prompt_RAG}.

Responses generated by the LLMs during the evaluation were manually assessed by an NLP researcher. The evaluation involved a binary judgment of responses as either correct or incorrect based on two main criteria: 1) \textbf{Content Faithfulness:} The response must accurately answer the question without any factual errors, 2) \textbf{Response Appropriateness:} If the question included specific instructions (e.g., “provide only one example”), the response is considered correct only if it follows these instructions.

The evaluator considered a response correct if it met both criteria.
Manual evaluation was preferred over automatic metrics (e.g., BLEU~\citep{papineni2002bleu}, METEOR~\citep{banerjee-lavie-2005-meteor}, BERTScore~\citep{DBLP:conf/iclr/ZhangKWWA20}) because these metrics do not adequately assess factual correctness, which is crucial in the business domain.
Additionally, redundant parts of responses, such as repeated sentences, were disregarded, focusing only on unique content.

\noindent
\textbf{lm-evaluation-harness:} The lm-evaluation-harness framework~\footnote{\url{https://github.com/Stability-AI/lm-evaluation-harness}} is a well-established tool for evaluating Japanese LLMs. It includes eight Japanese NLP tasks spanning various domains of language comprehension and generation. These tasks include reading comprehension (JSQuAD~\citep{kurihara-etal-2022-jglue}), QA (JCommonsenseQA~\citep{kurihara-etal-2022-jglue} and JAQKET~\footnote{\url{https://www.nlp.ecei.tohoku.ac.jp/projects/jaqket/}}), Natural language inference (JNLI~\citep{kurihara-etal-2022-jglue}), Summarization (XLSum-ja~\footnote{\url{https://huggingface.co/datasets/mkshing/xlsum_ja}}), Co-reference resolution (WinoGrande-ja~\citep{Keigo_Takahashi2023}), and Math problems (MGSM~\citep{zhang2023multicot}).

Due to the unavailability of the MARC-ja dataset~\citep{kurihara-etal-2022-jglue}, we conducted evaluations on the remaining seven tasks. Given that LLM performance can be significantly influenced by the prompts or templates used, we evaluated all available templates for each task and reported the highest score.

\subsection{Compared Models and Instruction Tuning}

We compare our pretrained model with six existing Japanese LLMs and three multilingual LLMs. The Japanese LLMs are categorized into two groups: models pretrained from scratch and models that have undergone continual pretraining from multilingual models.

Among the models pretrained from scratch, we compare three: 1) \textbf{llm-jp-13b}\citep{llmjp2024llmjpcrossorganizationalprojectresearch}, which is pretrained on Japanese Wikipedia, mC4, English Wikipedia, The Pile~\citep{gao2020pile}, and The Stack~\citep{kocetkov2022stack}; 2) \textbf{plamo-13b}\footnote{\url{https://huggingface.co/pfnet/plamo-13b}}, which is pretrained on English texts from RedPajama and Japanese texts from mC4 and Japanese Wikipedia; and 3) \textbf{weblab-10b}\footnote{\url{https://huggingface.co/matsuo-lab/weblab-10b}}, which is pretrained on Japanese texts from mC4 and English texts from The Pile. The first two models have 13 billion parameters, while the last has 10 billion.

We also include models that were continually pretrained from existing multilingual models: 4) \textbf{nekomata-14b}\citep{sawada2024release}, which is trained from qwen-14b on Japanese CC100, C4, Wikipedia, Oscar, Rinna curated corpus, and The Pile; 5) \textbf{ELYZA-japanese-13b}\footnote{\url{https://huggingface.co/elyza/ELYZA-japanese-Llama-2-13b}}, which is trained from Llama-2-13b on Japanese texts from Oscar and Wikipedia; and 6) \textbf{Swallow-13b}\citep{fujii2024continual}, which is trained from Llama-2-13b on Japanese texts from Wikipedia, meticulously cleaned Common Crawl, and English texts from RefinedWeb~\citep{penedo2023refinedweb} and The Pile. Models 4) has 14 billion parameters, while the others have 13 billion.

Additionally, we compare our models with multilingual LLMs: 7) \textbf{gpt-3.5-turbo-0125}, 8) \textbf{gpt-4-1106-preview}, selected from the OpenAI API; and 9) \textbf{Llama-2-13b-hf}~\citep{touvron2023llama}\footnote{\url{https://huggingface.co/elyza/ELYZA-japanese-Llama-2-13b}}, a 13-billion parameter English LLM capable of generating Japanese text.

LLMs that are not instruction-tuned often struggle to provide accurate answers, highlighting the importance of instruction-tuning. We evaluated several datasets for instruction-tuning in Japanese, including Ichikara~\citep{Ichikara}, Alpaca~\citep{alpaca_en, alpaca_ja}, Dolly~\citep{dolly_en, dolly_ja}, and OASST~\citep{köpf2023openassistant, oasst_ja}. Statistics for these datasets are provided in Table \ref{tab:dataset_explanation}.
Preliminary experiments shown in Table \ref{tab:compare_instruct_tune} reveal that the use of Ichikara dataset consistently performs well on question-answering tasks, with scores of 0.78 on JSQuAD, 0.86 on JAQKET, and 0.84 on JCommonsenseQA.
Therefore, we chose the Ichikara dataset for instruction-tuning our LLMs.
We use Low-Rank Adaptation (LoRA)~\citep{hu2021lowrank} for the instruction tuning.


\begin{table*}[t]
    \centering
    \small
    \begin{tabular}{l|lllllll|l}
    \hline
model & 
JSQUaD
& JAQKET 
& JCom.
& XW.
& JNLI
& MGSM 
& XLSum 
& Ave.\\ \hline
Ichikara &\textbf{0.78}&0.85&\textbf{0.84}&0.75&0.49&0.08&0.08&\textbf{0.550} \\ 
Alpaca& 0.73&0.85&0.81&0.73&\textbf{0.57}&0.07&0.07&0.545 \\
Dolly& 0.77&\textbf{0.87}&0.81&0.73&0.53&0.08&0.08&0.547 \\
OASST& 0.77&0.85&0.81&0.74&0.25&0.07&0.08&0.510 \\
\hline
    \end{tabular}
    \caption{Preliminary experiments: the comparison between datasets for instruction tuning. All base-model is our proposed 13-billion Japanese business domain specific LLM. JCom. and XW. refer to the JCommonsenseQA and XWinograd datasets, respectively. The Ave. column shows the averaged values over all datasets.}
    \label{tab:compare_instruct_tune}
\end{table*}

\section{Results}

\subsection{Results on Business Question Benchmark:} Table \ref{tab:result_business} displays the results from our domain-specific benchmark, categorizing the models into three groups: Japanese LLMs pretrained from scratch, Japanese LLMs with continual pretraining, and multilingual LLMs.

In the \textbf{NoContext-QA} setting, our pretrained model achieves the highest accuracy of $0.90$. This is notably higher compared to other Japanese LLMs such as nekomata-14b and Swallow-13b, which both score $0.78$. This performance indicates that our model effectively utilizes domain-specific knowledge stored internally without relying on external documents.

For the \textbf{ManualRAG-QA} task, our model again performs best among LLMs pretrained from scratch, achieving an accuracy of $0.84$.
In contrast, the other models in this category score $0.76$ (llm-jp-13b), $0.82$ (plamo-13b), and $0.52$ (weblab-10b).
A similar trend is observed in the \textbf{AutoRAG-QA} task, where our model scores $0.74$, while other models score $0.62$ (llm-jp-13b), $0.64$ (plamo-13b), and $0.34$ (weblab-10b), respectively.
In these RAG settings, which require comprehension and the ability to extract information from documents, our models perform better than other full-scratch LLMs.
However, Japanese LLMs with continual pretraining, such as ELYZA-japanese-13b and Swallow-13b, achieve higher scores of $0.94$ and $0.90$ in \textbf{ManualRAG-QA}, respectively.
These values suggest that continual pretraining works better in comprehension.

When comparing pretraining strategies, continual pretraining outperforms full-scratch models except for our model.
For instance, in the \textbf{NoContext-QA} task, models with continual pretraining, such as Swallow-13b and nekomata-14b, achieve scores of $0.78$, whereas the best pretrained full-scratch baseline, llm-jp-13b, scores $0.34$. Despite the overall advantage of continual pretraining, our model performs the best in the \textbf{NoContext-QA} task, highlighting the value of domain-specific data to strengthen the internal domain-specific knowledge in LLMs.

For tasks involving context, such as \textbf{ManualRAG-QA} and \textbf{AutoRAG-QA}, our model achieves the highest scores of $0.84$ and $0.74$, respectively, among full-scratch models.
However, models with continual pretraining outperform in these tasks with scores of $0.94$ and $0.82$. This may be attributed to the use of Llama2 as a base model for continual pretraining, which benefits from being pretrained on a larger dataset and thus has stronger language comprehension capabilities.

Although this study focuses on models pretrained from scratch, future work could explore the potential performance gains from continually training a strong existing model, such as Llama2, on a domain-specific corpus.

\begin{table*}
    \centering
    \begin{tabular}{l|ccc}
    \hline
model & {\bf NoContext-QA} & {\bf ManualRAG-QA} & {\bf AutoRAG-QA} \\ \hline
\multicolumn{4}{l}{Japanese LLMs trained from scratch} \\ \hline
-  \   Our model	& \textbf{0.90} & \textbf{0.84} & \textbf{0.74}\\ 
- 1) llm-jp-13b & 0.34 & 0.76 & 0.62\\ 
- 2) plamo-13b	& 0.34 & 0.82 & 0.64 \\
- 3) weblab-10b & 0.26 & 0.52 & 0.34 \\ \hline
\multicolumn{4}{l}{Japanese LLMs with continual pretraining} \\ \hline
- 4) nekomata-14b & \textbf{0.78} & 0.74 & 0.76\\
- 5) ELYZA-japanese-13b & 0.32 & \textbf{0.94} & 0.70 \\
- 6) Swallow-13b & \textbf{0.78} & 0.90 & \textbf{0.82}\\ \hline
\multicolumn{4}{l}{Multilingual LLMs} \\ \hline
- 7) gpt-3.5-turbo-0125 & 0.54 & 0.62 & 0.34\\
- 8) gpt-4-1106-preview & \textbf{0.78} & \textbf{0.94} & \textbf{0.86}\\
- 9) Llama-2-13b-hf & 0.24 & 0.84 & 0.64\\
\hline
    \end{tabular}
    \caption{Results on business-specific benchmark. Each score stands for the accuracy. The values obtained from the best performing models in each category is shown in bold.}
    \label{tab:result_business}
\end{table*}

\subsection{Results on General-domain Benchmark: }Table \ref{tab:result_instruct} presents the results on the general-domain benchmark, ``lm-evaluation-harness.''
Among the models pretrained from scratch, our model demonstrates superior performance in terms of the averaged score i.e., ours achieves $0.55$ on average while the best performing Japanese LLMs pretrained from scratch achieve $0.49$.
In particular, in QA tasks without context, such as JAQKET and JCommonsenseQA, our model achieves the best scores $0.78$ and $0.85$, respectively.

\begin{table*}[t]
    \centering
    \small
    \begin{tabular}{l|ccccccc|c}
    \hline
model & 
JSQuAD
& JAQKET 
& \begin{tabular}{l}JCom.\end{tabular} 
& XW.
& JNLI
& MGSM 
& XLSum 
& Ave.\\ \hline
\multicolumn{9}{l}{Japanese LLMs Pretrained from Scratch} \\ \hline
- Ours& \textbf{0.78}&\textbf{0.85}&\textbf{0.84}&\textbf{0.75}&\textbf{0.49}&\textbf{0.08}&0.08&\textbf{0.55} \\
- 1) weblab-10b& 0.72&0.43&0.65&0.67&0.30&0.02&0.05&0.41  \\
- 2) plamo-13b&  0.68&0.69&0.64&0.68&0.41&0.02&\textbf{0.10}&0.46  \\
- 3) llm-jp-13b&  0.69&0.76&0.79&0.70&0.37&0.02&\textbf{0.10}&0.49   \\\hline
\multicolumn{9}{l}{Japanese LLMs with Continual Pretraining} \\ \hline
- 4) nekomata-14b& \textbf{0.87}&0.88&\textbf{0.94}&\textbf{0.80}&\textbf{0.65}&\textbf{0.36}&\textbf{0.22}&\textbf{0.68} \\ 
- 5) ELYZA-13b& 0.79&0.75&0.87&0.78&0.51&0.10&0.19&0.57  \\ 
- 6) Swallow-13b& 0.86&\textbf{0.91}&0.91&0.72&0.52&0.18&0.20&0.61  \\ \hline
\multicolumn{9}{l}{Multilingual LLMs pretrained from Scratch} \\ \hline
- 9) Llama-2-13b-hf& 0.81&0.75&0.82&0.63&0.47&0.12&0.21&0.54  \\ \hline
    \end{tabular}
    \caption{Results on lm-evaluation-harness. JCom. and XW. refer to the JCommonsenseQA and XWinograd datasets, respectively. The Ave. column shows the averaged values over all datasets. OpenAI's GPTs cannot be used for this experiments because the model parameters are not public.}
    \label{tab:result_instruct}
\end{table*}


\subsection{Analysis of Updated Models} We compare the updated model with the original pretrained model.
The left part of Table \ref{tab:result_continuous} shows the validation loss values on two datasets: the 8,192 documents that were used for the validation portion during pretraining (Pretrain Data) and the latest business documents used for updating the model (Latest Data).
If our LLM were able to learn the latest knowledge without losing general knowledge, it would achieve low validation loss on both Pretrain Data and Latest Data.
The loss value for the updated model with $r$ = 0.0, indicating that it was trained solely on the latest documents, is 2.05 on Latest Data.
This represents an improvement over the original pretrained model’s loss of 2.25, suggesting that the updated model acquired the latest knowledge better. However, the loss on the Pretrain Data increased from 2.11 to 2.19 compared to the original model, which suggests the worse fit to the non-latest documents.

Increasing the value of $r$, which incorporates more pretraining data along with the latest documents, mitigates the increase in loss on the pretraining data.
For example, with  $r$ set to 0.3, the loss is 2.12 on Pretrain Data, which is nearly equivalent to the original model’s loss of 2.11, while the loss on the latest documents remains stable at 2.04.

The right part of Table \ref{tab:result_continuous} shows the accuracy of our compared updated LLMs in question answering about latest news.
We created a set of 10 questions (LatestQ) about recent business topics from October to November 2023.
We selected topics that had shown a notable increase in search engine access compared to September 2023.
As result, all the updated models achieve a higher accuracy (0.90) compared to the pretrained model (0.30), indicating that a Japanese- and business-specific model can acquire new information through continual learning.
Unlike the results regarding loss, we did not observe significant differences in accuracy among models with different values of $r$ on the Business Question Benchmark (Non-LatestQ) where we obtained high values e.g., $0.90$ (Pretrained model and Updated model($r=0.3$)) or $0.92$ (Updated model ($r=0.0,0.1$)).
We conclude that using only the latest articles can lead to degradation in both loss and accuracy and incorporating around 10\% of non-latest articles proves effective, while including 30\% is excessive.

\begin{table*}[t]
    \centering
    \begin{tabular}{l|cc|cc}
    \hline
& \multicolumn{2}{c|}{Loss $\downarrow$} 
& \multicolumn{2}{c}{Accuracy $\uparrow$} \\ \hline
& Pretrain Data
& Latest Data
& Non-LatestQ. & LatestQ.\\ \hline 
Pretrained model & \textbf{2.11}& 2.25 & 0.90 & 0.30\\
Updated model ($r = 0.0$)	& 2.19& \textbf{2.05} & \textbf{0.92} & \textbf{0.90} \\
Updated model ($r = 0.1$)	& 2.14& \textbf{2.05} & \textbf{0.92} & \textbf{0.90}\\
Updated model ($r = 0.3$)	& 2.12& \textbf{2.04} & 0.90 & \textbf{0.90} \\ \hline
    \end{tabular}
    \caption{Comparisons of validation losses on the pretrain data and the latest data, and accuracies on the questions about the latest business  and non-latest business knowledge (Latest Q. and Non-latest Q.).}
    \label{tab:result_continuous}
\end{table*}



\subsection{Examples of Outputs}
Table \ref{tab:examples_output} shows the examples of the outputs from our pretrained model and compared LLMs.
When we input a question ``Which two banks failed in March 2023?'', our pretrained model correctly generates two banks: the Silicon Valley Bank and the Signature Bank.
Other three Japanese LLMs' outputs contain hallucinations, which are indicated in bold in the table.
For example, Wells Forgo, Bank of America, and Citigroup did not fail in March 2023.
We observed a similar tendency for the other questions.

\begin{table*}
    \centering
    \footnotesize
\begin{tabularx}{\linewidth}{lX}
\toprule
\hline
\multicolumn{2}{l}{Question : Which two American banks went bankrupt in March 2023?}\\
\hline
Model & Generated Text \\\hline
\midrule
Ours & The two U.S. banks that failed in March 2023 were Silicon Valley Bank and Signature Bank. (\begin{CJK}{UTF8}{ipxg}2023年3月に経営破綻した米国の2つの銀行は、シリコンバレー銀行とシグネチャー銀行である。)\end{CJK})\\\hline
\midrule
llm-jp-13b-v1.0 & The two U.S. banks that failed in March 2023 were {\bf Wells Fargo and Bank of America}. (\begin{CJK}{UTF8}{ipxg}2023年3月に経営破綻したアメリカの2つの銀行は、\bf{ウェルズ・ファーゴとバンク・オブ・アメリカである。}\end{CJK})  \\\hline
\midrule
plamo-13b & The two banks that failed in March 2023 were {\bf Bank of America and Citigroup}. (\begin{CJK}{UTF8}{ipxg}2023年3月に経営破綻した2つの銀行は、{\bf バンク・オブ・アメリカとシティ・グループ}でした。\end{CJK})\\\hline
\midrule
weblab-10b & The two U.S. banks that failed in March 2023 were {\bf Bank of America and Wells Fargo}. (\begin{CJK}{UTF8}{ipxg}2023年3月に経営破綻したアメリカの2つの銀行は、Bank of America と Wells Fargoです。\end{CJK}) \\\hline
\midrule
ChatGPT (gpt-3.5-turbo-0125) & The two American banks that went bankrupt in March 2023 are \bf{Regional Asahi Bank and Centry National Bank}. (\begin{CJK}{UTF8}{ipxg}2023年3月に経営破綻したアメリカの2つの銀行は、リージョナル朝陽銀行とセントリーナショナル銀行です。\end{CJK})\\\hline
\bottomrule
\end{tabularx}

\caption{Translated examples of output generated from different LLMs. The sentences in the buckets are the original outputs in Japanese. Hallucinated terms are presented in bold.}
\label{tab:examples_output}
\end{table*}
\section{Related Work}
Recent major LLMs are multilingual models such as Llama2~\citep{touvron2023llama}, OpenAI's GPTs~\citep{openai}, and BLOOM~\citep{workshop2023bloom}.
The datasets used for such major models often include a high percentage of English data.
Therefore, performances in languages other than English have been still underexplored.
Studies targeting non-English languages, such as Chinese~\citep{zhang2021cpm}, German~\citep{german_gpt}, and Japanese~\citep{swallow}, increase the proportion of non-English texts used in pretraining.
Whereas, we propose an approach to build a large size of Japanese specific corpus by filtering out other langauges with language identification libraries and a noise character detection, and we pretrain Japanese-specific LLM. 

The language-specific datasets for instruction tuning have been released.
For Japanese, Ichikara, Alpaca, Dolly, and OASST are common; however, the performances of models trained on these datasets have not been compared in depth.
Our preliminary experiment is the first to compare these datasets, which can be considered as an important contribution for the Japanese LLM community.
Also, our experiments can provide insights for researchers who focus on other non-English languages.

Domain-specific models are pretrained in two different ways; training from scratch or continual training.
Representative examples of the former started for the encoder-only models such as SciBERT~\citep{beltagy-etal-2019-scibert} in the science domain and later decoder-only large language models follow, e.g., BloombergGPT~\citep{wu2023bloomberggpt}.
The continually learned domain-specific LMs include BioBERT~\citep{lee2020biobert}, exBERT~\citep{wang-etal-2020-extending}.
Scratch approaches are often used for settings where we can get sufficient data.
Our setting is categorized into this setting, thus, we use the scratch approach.

The combination of the two directions, i.e., language- and domain-specific settings, is underexplored in Japanese.
Japanese domain-specific language model does not exist, except for the pretrained BERT for the material science domain~\citep{ishigaki2023pretraining}.
Our target is decoder-only architecture with many more parameters for the business domain, which has high demand in the industry but is less studied.
Our pretrained model is the first Llama-based domain-specific model for Japanese.

Continual pretraining is a promising direction if we have only a small dataset for pretraining.
Our experiments for updating the pretrained model with the latest news are categorized in this setting.
``Catastrophic forgetting'' is a major problem in this case~\citep{ling2023domain}.
\citet{scialom-etal-2022-fine} suggest that the mixture of two types of data can mitigate this problem, thus, we used the technique to mix the latest documents and older ones.
\section{Conclusion}
This paper presented the first Japanese business domain-specific LLM.
We pretrain the LLM with 13 billion parameters from scratch and the model is released to be publicly available.
We also update the model parameters by the latest articles and confirmed performance gains.
For updates, we conclude that using only the latest articles can lead to performance degradation but incorporating around 10\% of non-latest articles proves effective.
Comprehensive evaluations demonstrated that our LLM achieves the best accuracy score without retrieval on the domain-specific benchmark we newly released.
These results provide valuable insights for researchers working on other domains and languages.
For future work, we will explore comparing different ways to train language- and domain-specific LLMs e.g., full-stratch v.s. continual pretraining.


\section{Ethics Statement}
The proposed model is still in the early stages of research and development, and its output has not yet been adjusted to align with safety considerations. However, adjustments will be made in the future to ensure that the model takes safety into account.
\newpage

\bibliography{paper}

\newpage

\appendix
\section{Appendix}
\subsection{Templates used in Experiments}

Table \ref{tab:templete_used} shows the versions of templates used in our experiments.

\begin{table*}[h]
    \centering
\begin{tabularx}{\linewidth}{cccc}
\toprule
Task & Task Version & Prompt Version & Number of Few-shot \\
\midrule
JSQuAD & 1.1 & 0.1, 0.2, 0.3 & 2 \\
\midrule
JAQKET & v2-0.2 & 0.1, 0.2, 0.3 & 1 \\
\midrule
JCommonsenseQA & 1.1 & 0.1, 0.2.1, 0.3 & 3 \\
\midrule
JWinograd & ja & - & 0 \\
\midrule
JNLI & 1.3 & 0.2, 0.3 & 3 \\
\midrule
MGSM & 1.0 & 0.0, 0.3 & 5 \\
\midrule
XLSum & ja-1.0 & 0.0, 0.3 & 1 \\
\bottomrule
\end{tabularx}
\caption{Settings of the template of the lm-evaluation-harness. We experimented every model in Table \ref{tab:result_instruct} with the templates listed here. The scores reported in Table \ref{tab:result_instruct} are the highest among the variants of the templates.}
\label{tab:templete_used}
\end{table*}

\subsection{Prompts used in Experiments}
\label{apx:prompts}

\begin{figure}[h]
\small
\setbox0\vbox{
\vbox{Please answer the question briefly.
\\\\
Question:\{question\}
\\\\
\#\#\# Output:
}
}
\centerline{\fbox{\box0}}
\caption{Translated prompt for NoContext-QA. In the experiment, Japanese prompt was used.}
\label{fig:prompt_NoContext}
\end{figure}

\begin{figure}[h]
\small
\setbox0\vbox{
\vbox{Please answer to the given question. If the answer to the question is included in the article text, please use the answer from the text. If the article does not contain the answer please state that "the article does not contain the answer" and answer the question using your knowledge.
\\\\
Question:\{question\}
\\\\
Article Text:
\{the first 1k characters\}
\\\\
\#\#\# Output:
}
}
\centerline{\fbox{\box0}}
\caption{Translated prompt for ManualRAG-QA and AutoRAG-QA. In the experiment, Japanese prompt was used.}
\label{fig:prompt_RAG}
\end{figure}



\end{document}